\documentclass{article}

\usepackage[utf8]{inputenc} 
\usepackage[T1]{fontenc}    
\usepackage{hyperref}       
\usepackage{url}            
\usepackage{booktabs}       
\usepackage{amsfonts}       
\usepackage{nicefrac}       
\usepackage{microtype}      
\usepackage{amsmath}
\usepackage{graphicx}
\usepackage{subcaption}
\usepackage{comment}
\usepackage{tabu}
\usepackage[accepted]{icml2018}

\def\bb{{\boldsymbol b}}

\def\bh{{\boldsymbol h}}

\def\bx{{\boldsymbol x}}

\def\bz{{\boldsymbol z}}

\def\bd{{\boldsymbol d}}

\def\R{{\mathbb{R}}}
\def\N{{\mathcal{N}}}
\def\L{{\mathcal{L}}}
\def\E{{\mathbb{E}}}

\definecolor{darkblue}{rgb}{0.0, 0.0, 0.55}
\hypersetup{
    colorlinks=true,
    citecolor=darkblue
}

\begin{document}

\twocolumn[
\icmltitle{Stochastic WaveNet: A Generative Latent Variable Model for Sequential Data}

\icmlsetsymbol{equal}{*}

\begin{icmlauthorlist}
\icmlauthor{Guokun Lai}{cmu}
\icmlauthor{Bohan Li}{cmu}
\icmlauthor{Guoqing Zheng}{cmu}
\icmlauthor{Yiming Yang}{cmu}
\end{icmlauthorlist}

\icmlaffiliation{cmu}{Language Technology Institute, 
Carnegie Mellon University, Pittsburgh, PA 15213, USA}

\icmlcorrespondingauthor{Guokun Lai}{guokun@cs.cmu.edu}

\icmlkeywords{Machine Learning, ICML}

\vskip 0.3in
]

\printAffiliationsAndNotice{} 

\begin{abstract}
How to model distribution of sequential data, including but not
limited to speech and human motions, is an important ongoing research
problem. It has been demonstrated that model capacity can be
significantly enhanced by introducing stochastic latent variables in
the hidden states of recurrent neural networks. Simultaneously,
WaveNet, equipped with dilated convolutions, achieves astonishing
empirical performance in natural speech generation task. In this
paper, we combine the ideas from both stochastic latent variables and
dilated convolutions, and propose a new architecture to model
sequential data, termed as Stochastic WaveNet, where stochastic latent
variables are injected into the WaveNet structure. We argue that
Stochastic WaveNet enjoys powerful distribution modeling capacity and
the advantage of parallel training from dilated convolutions. In order
to efficiently infer the posterior distribution of the latent
variables, a novel inference network structure is designed based on 
the characteristics of WaveNet architecture. State-of-the-art performances
on benchmark datasets are obtained by Stochastic WaveNet on natural
speech modeling and high quality human handwriting samples can be
generated as well.
\end{abstract}

\section{Introduction}

Learning to capture complex distribution of sequential data is an important
machine learning problem and has been extensively studied in recent
years. The autoregressive neural network models, including Recurrent
Neural Network \citep{hochreiter1997long, chung2014empirical},
PixelCNN \citep{oord2016pixel} and WaveNet \citep{van2016wavenet}, have
shown strong empirical performance in modeling natural language,
images and human speeches. All these methods are aimed at learning a
deterministic mapping from the data input to the output. Recently,
evidence has been found \citep{fabius2014variational, gan2015deep,
  gu2015neural, goyal2017z, shabanian2017variational} that probabilistic
modeling with neural networks can benefit from uncertainty introduced to
their hidden states, namely including stochastic latent variables in
the network architecture. Without such uncertainty in the hidden
states, RNN, PixelCNN and WaveNet would parameterize the
randomness only in the final layer by shaping a output distribution from the specific distribution family. Hence the output distribution (which
is often assumed to be Gaussian for
continuous data) would be unimodal or the mixture of unimodals given the
input data, which may be insufficient to capture the complex true data
distribution and to describe the complex correlations among different
output dimensions \citep{boulanger2012modeling}. Even for the non-parametrized discrete output distribution modeled by the softmax function, a phenomenon referred to as softmax bottleneck \citep{yang2017breaking} still limits the family of output distributions. By injecting the stochastic latent variables into the hidden states and transforming their uncertainty to outputs by non-linear layers, the stochastic neural network is equipped with the ability to model the data with a much richer family of distributions.

Motivated by this, numerous variants of RNN-based stochastic neural
network have been proposed. STORN \citep{bayer2014learning} is the
first to integrate stochastic latent variables into RNN's hidden
states. In VRNN \citep{chung2015recurrent}, the prior of stochastic
latent variables is assumed to be a function over historical data and stochastic latent variables,
which allows them to capture temporal dependencies. SRNN
\citep{fraccaro2016sequential} and Z-forcing \citep{goyal2017z} offer more powerful versions with augmented inference networks which better capture the correlation between the stochastic latent variables and the whole observed sequence.
By introducing stochasticity to the hidden states, these RNN-based
models achieved significant improvements over vanilla RNN models on
log-likelihood evaluations on multiple benchmark datasets from various
domains \citep{goyal2017z,shabanian2017variational}.

In parallel with RNN, WaveNet \citep{van2016wavenet} provides another
powerful way of modeling sequential data with dilated convolutions,
especially in the natural speech generation task. While RNN-based
models must be trained in a sequential manner, training a WaveNet can
be easily parallelized. Furthermore, the parallel WaveNet proposed in
\citep{oord2017parallel} is able to generate new sequences in
parallel. WaveNet, or dilated convolutions, has also been adopted as
the encoder or decoder in the VAE framework and produces reasonable
results in the text \citep{semeniuta2017hybrid,
  yang2017improved} and music \citep{engel2017neural} generation task. 

In light of the advantage of introducing stochastic latent variables
to RNN-based models, it is natural to raise a problem whether this benefit carries
to WaveNet-based models. To this end, in this paper we propose
Stochastic WaveNet, which associates stochastic latent variables with
every hidden states in the WaveNet architecture. Compared with the
vanilla WaveNet, Stochastic WaveNet is able to capture a richer family
of data distributions via the added stochastic latent variables. It
also inherits the ease of parallel training with dilated convolutions
from the WaveNet architecture. Because of the added stochastic latent
variables, an inference network is also designed and trained jointly
with Stochastic WaveNet to maximize the data log-likelihood. We
believe that after model training, the multi-layer structure of latent
variables leads them to reflect both hierarchical and sequential
structures of the data. This hypothesis is validated empirically by controlling the number of layers of stochastic latent
variables.

The rest of this paper is organized as follows: we briefly
review the background in Section \ref{sec:preliminary}. The proposed
model and optimization algorithm are introduced in Section
\ref{sec:model}. We evaluate and analyze the proposed model on
multiple benchmark datasets in Section \ref{sec:experiment}. Finally,
the summary of this paper is included in Section \ref{sec:conclusion}.

\section{Preliminary}
\label{sec:preliminary}

\subsection{Notation}

We first define the mathematical symbols used in the rest of this paper. We denote a set of vectors by a bold symbol, such as $\bx$, which may utilize one or two dimension subscripts as index, such as $\bx_i$ or $\bx_{i,j}$. $f(\cdot)$ represents the general function that transforms an input vector to a output vector. And $f_\theta(\cdot)$ is a neural network function parametrized by $\theta$. For a sequential data sample $\bx$, $T$ represents its length. 

\subsection{Autoregressive Neural Network}

Autoregressive network model is designed to model the joint distribution of the high-dimensional data with sequential structure, by factorizing the joint distribution of a data sample as
\begin{equation}
p(\bx) = \prod_{t=1}^T p_\theta(x_t|x_{<t})
\end{equation}
where $\bx = \{x_1, x_2, \cdots x_T\}, x_t \in R^{d}$, $t$ indexes the temporal time stamps, and $\theta$ represents the model parameters. Then the autoregressive model can compute the likelihood of a sample and generate a new data sample in a sequential manner. 

In order to capture richer stochasticities of the sequential generation process, stochastic latent variables for each time stamp have been introduced, referred to as stochastic neural network \citep{chung2015recurrent,fraccaro2016sequential,goyal2017z}. Then the joint distribution of the data together with the latent variables is factorized as,

\begin{equation}
\begin{aligned}
p(\bx, \bz) &= \prod_{t=1}^T p_\theta(x_t, z_t| x_{<t}, z_{<t}) \\
&=\prod_{t=1}^T p_\theta(x_t| x_{<t}, z_{\le t}) p_\theta(z_t|x_{<t}, z_{<t})
\end{aligned}
\end{equation}

where $\bz= \{z_1, z_2, \cdots z_T\}, z_t \in R^{d'}$ has the same sequence length as the data sample,  $d'$ is its dimension for one time stamp. $\bz$ is also generated sequentially, namely the prior of $z_t$ is conditional probability given $x_{<t}$ and $z_{<t}$.

\subsection{WaveNet}

\begin{figure}[ht]
\centering
\includegraphics[trim={6.625in 0 4.95in 0},clip,width=0.85\linewidth]{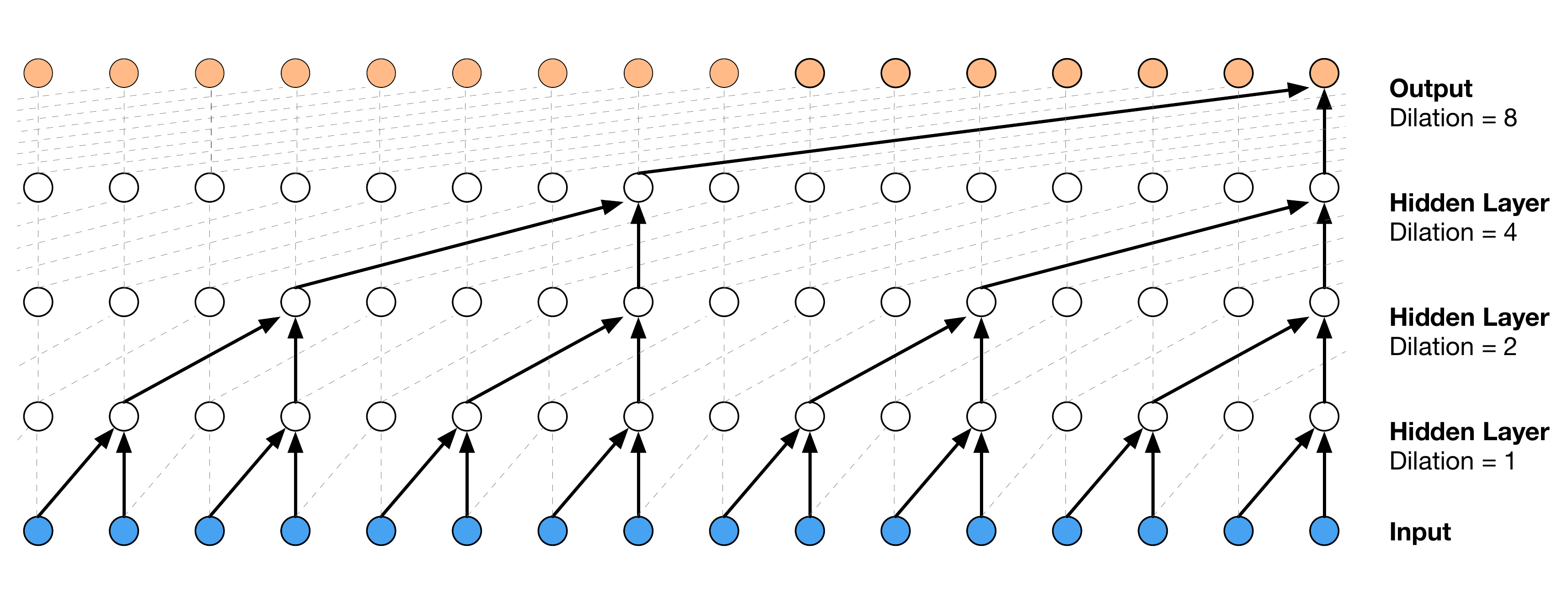}
\caption{Visualization of a WaveNet structure from \citep{van2016wavenet}}
\label{fig:wavenet}
\end{figure}

WaveNet \citep{van2016wavenet} is a convolutional autoregressive neural network which adopts dilated causal convolutions \citep{yu2015multi} to extract the sequential dependency in the data distribution. Different from recurrent neural network, dilated convolution layers can be computed in parallel during the training process, which makes WaveNet much faster than RNN in modeling sequential data. A typical WaveNet structure is visualized in Figure \ref{fig:wavenet}. Beside the computation advantage, WaveNet has shown the start-of-the-art result in speech generation task \citep{oord2017parallel}.

\section{Stochastic WaveNet}

\label{sec:model}

In this section, we introduce a sequential generative model
(Stochastic WaveNet), which imposes stochastic latent variables with
the multi-layer dilated convolution structure. We firstly introduce
the generation process of Stochastic WaveNet, and then describe the
variational inference method.

\subsection{Generative Model}

Similar as stochastic recurrent neural networks, we inject the stochastic latent variable in each WaveNet hidden node in the generation process, which is illustrated in Figure \ref{fig:generation}. More specifically, for a sequential data sample $\bx$ with length $T$, we introduce a set of stochastic latent variables $\bz = \{z_{t,l}|1 \le t \le T, 1 \le l \le L\},  z_{t,l} \in \R^{d'}$,  where $L$ is the number of the layers of WaveNet architecture. Then the generation process can be described as,

\begin{equation}
\begin{aligned}
p(\bx, \bz) &= \prod_{t=1}^T p_\theta(x_t, z_{t,1:L}| x_{<t}, z_{<t, 1:L}) \\
&= \prod_{t=1}^T [p_\theta(x_t | z_{\le t,1:L}, x_{<t}) \\
&\qquad \quad \prod_{l=1}^L p_\theta(z_{t, l}| z_{t, <l}, x_{<t}, z_{<t, 1:L})]
\label{eq:pxz}
\end{aligned}
\end{equation}

The generation process can be interpreted as this. At each time stamp $t$,  we sample the stochastic latent variables $z_{t,l}$ from a prior distribution which are conditioned on the lower level latent variables and historical records including the data samples $x_{<t}$ and latent variables $z_{<t}$. Then we sample the new data sample $x_{t}$ according to all sampled latent variables and historical records. Through this process, new sequential data samples are generated in a recursive way. 

In Stochastic WaveNet, the prior distribution $p_\theta(z_{t, l}| x_{<t}, z_{t, <l}, z_{<t, 1:L}) = \N (z_{t,l}; \mu_{t,l}, v_{t,l})$ is defined as a Gaussian distribution with the diagonal covariance matrix. The sequential and hierarchical dependency among the latent variables are modeled by the WaveNet architecture. In order to summarize all historical information, we introduce two stochastic hidden variables $\bh$ and $\bd$, which are calculated as,

\begin{equation}
\begin{aligned}
h_{t,l} &= f_{\theta_1}(d_{t - 2^l, l-1}, d_{t, l-1}) \\
d_{t,l} &= f_{\theta_2}(h_{t,l}, z_{t,l})
\end{aligned}
\label{eq:forward}
\end{equation}

Where $f_{\theta_1}$ mimics the design of the dilated convolution in WaveNet, and $f_{\theta_2}$ is a fully connected layer to summarize the hidden states and the sampled latent variable. Different from the vanilla WaveNet, the hidden states $\bh$ are \textit{stochastic} because of the random samples $\bz$. We parameterize the mean and variance of the prior distributions by the hidden representations $\bh$, which is $\mu_{t,l} = f_{\theta_3}(h_{t,l})$ and $\log~v_{t,l} = f_{\theta_4}(h_{t,l})$. Similarly, we parameterize the emission probability $p_\theta(x_t | x_{<t}, z_{t,1:L}, z_{<t, 1:L})$ as a neural network function over the hidden representations.

\subsection{Variational Inference for Stochastic WaveNet}

Instead of directly maximizing log-likelihood for a sequential sample $\bx$, we optimize its variational evidence lower bound (ELBO) \citep{jordan1999introduction}. Exact posterior inference of the stochastic variables $\bz$ of Stochastic WaveNet is intractable. Hence, we describe a variational inference method for Stochastic WaveNet by utilizing the reparameterization trick introduced in \citep{kingma2013auto}.  Firstly, we write the ELBO as,

\begin{equation}
\begin{aligned}
\log~p(\bx) & \ge \int q_\phi(\bz|\bx) \log \frac{p_\theta(\bx, \bz)}{q_\phi(\bz|\bx)} dz \\
& = \E_{q_\phi(\bz|\bx)}[\sum_{t=1}^T \log~p_\theta(x_t|z_{\le t, 1:L}, x_{<t})] \\
& \qquad - D_{KL}(q_\phi(\bz|\bx) || p_\theta(\bz|\bx) )\\
& = \L(x) \\
where&~~p_\theta(\bz|\bx) = \prod_{t=1}^T \prod_{l=1}^L p_\theta(z_{t, l}| z_{t, <l}, x_{<t}, z_{<t, 1:L}) \\
\end{aligned}
\end{equation}

We can derive the second equation by taking Eq. \ref{eq:pxz} into the first equation, and $\L(x)$ denotes the loss function for the sample $\bx$. Here another problem needs to be addressed is how to define the posterior distribution $q_\phi(\bz|\bx)$. In order to maximize the ELBO, we factorize the posterior as,

\begin{equation}
\begin{aligned}
q_\phi(\bz|\bx) = \prod_{t=1}^T \prod_{l=1}^L q_\phi(z_{t, l}| \bx, z_{t, <l}, z_{<t, 1:L}) 
\end{aligned}
\end{equation}

Here the posterior distribution for $z_{t,l}$ is conditioned on the stochastic latent variables sampled before it and the entire observed data $\bx$. By utilizing the future data $x_{\ge t}$, we can better maximize the first term in $\L(x)$, the reconstruction loss term. In opposite, the prior distribution of $z_{t,l}$ is only conditioned on $x_{<t}$, so encoding $x_{\ge t}$ information may increase the degree of distribution mismatch between the prior and posterior distribution, namely enlarging the KL term in loss function. 

{\bf{Exploring the dependency structure in WaveNet}.} However, by analyzing the dependency among the outputs and hidden states of WaveNet, we would find that the stochastic latent variables at time stamp t, $z_{t,1:L}$ would not influence whole posterior outputs. So the inference network would only require partial posterior outputs to maximize the reconstruction loss term in the loss function. Denote the set of outputs that would be influenced by $z_{l,t}$ as $s(l,t)$. The posterior distribution can be modified as,

\begin{equation}
\begin{aligned}
q_\phi(\bz|\bx) = \prod_{t=1}^T \prod_{l=1}^L q_\phi(z_{t, l}| x_{<t}, x_{s(l,t)}, z_{t, <l}, z_{<t, 1:L}) 
\end{aligned}
\end{equation}

\begin{figure}[t]
\centering
\begin{subfigure}{.37\textwidth}
  \includegraphics[width=\linewidth]{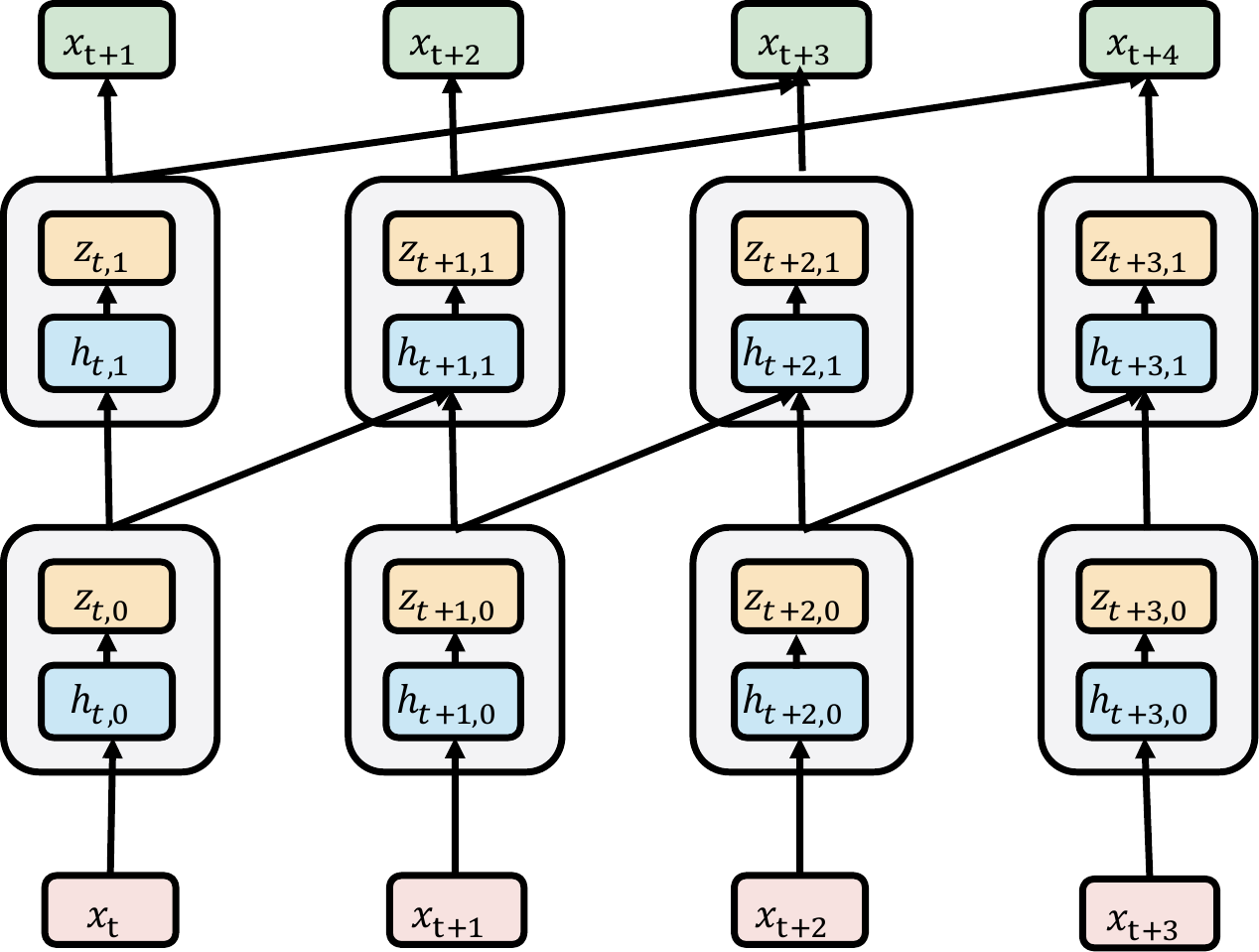}
  \caption{Generative Model}
  \label{fig:generation}
\end{subfigure}
\hspace{0.5cm}
\begin{subfigure}{.4\textwidth}
  \includegraphics[width=\linewidth]{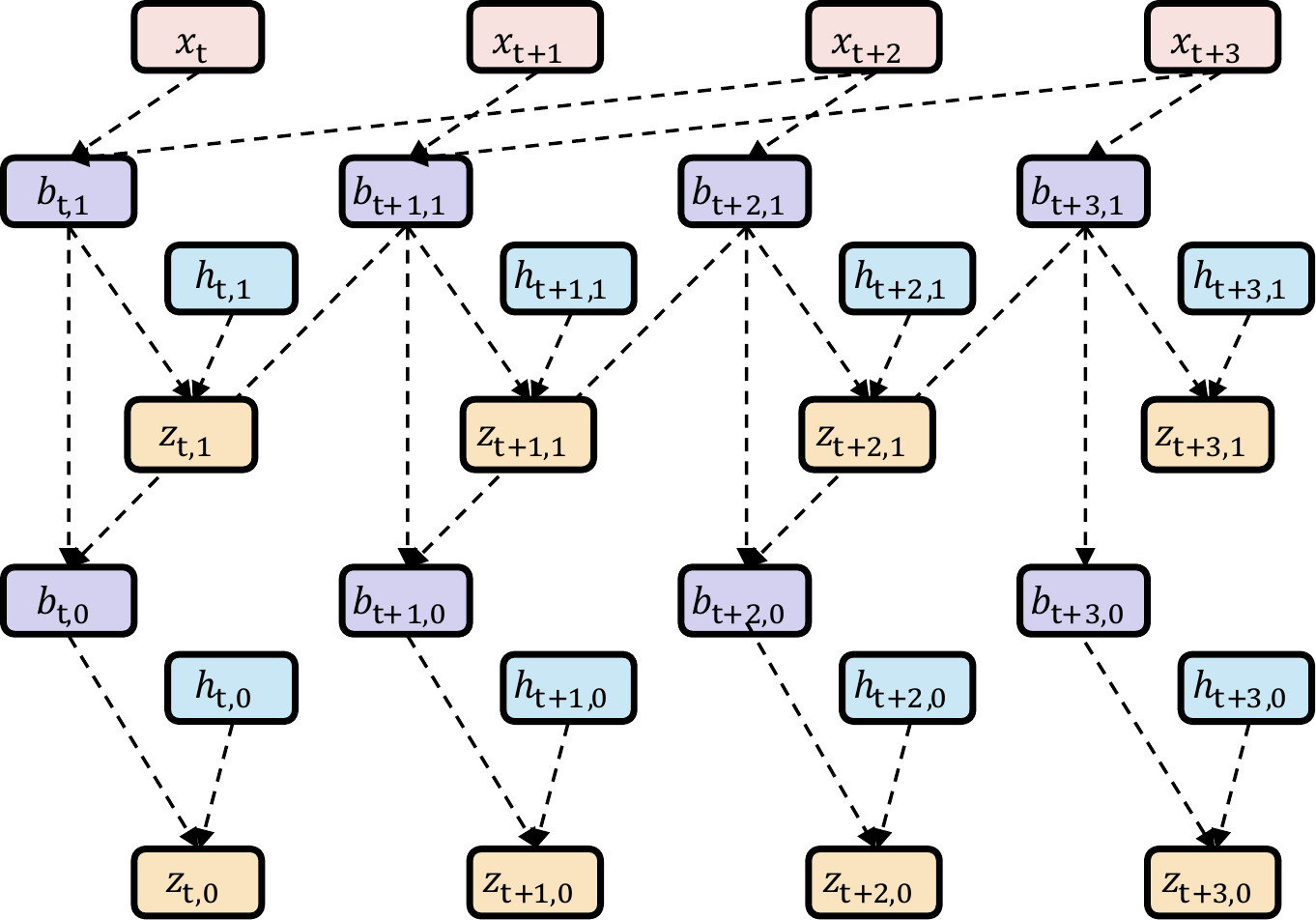}
  \caption{Inference Model}
  \label{fig:inference}
\end{subfigure}
\caption{This figure illustrates a toy sample of Stochastic WaveNet, which has two layers. The left one is the generative model, and the right one is the inference model. Both the solid line and dash line represent the neural network functions. The $\bh$ in the right figure is identical to the one in the left. The $\bz$ in the generative model are sampled from prior distributions, and the ones in the inference model are from posterior.}
\label{fig:SWAVE}
\end{figure}

The modified posterior distribution removes unnecessary conditional variables, which makes the optimization more efficient. To summarize the information from posterior outputs $x_{s(l,t)}$, we design a reversed WaveNet architecture to compute the hidden feature $\bb$, illustrated in Figure \ref{fig:inference}, and $b_{t,l}$ is formulated as, 

\begin{equation}
b_{t, l} = f(x_{s(t,l)}) = f_{\phi_1} (b_{t, l+1}, b_{t+2^{l+1}, l+1})
\end{equation}

where we define that $b_{t,L} = f_{\phi_2}(x_{t}, x_{t+2^{L+1}})$, and $f_{\phi_1}$ and $f_{\phi_2}$ is the dilated convolution layer, whose structure is a reverse version of $f_{\theta_1}$ in Eq.\ref{eq:pxz}. Finally, we inference the posterior distribution $q_\phi(z_{t, l}| \bx, z_{t, <l}, z_{<t, 1:L}) = \N(z_{t, l}; \mu'_{t,l}, v'_{t,l})$ by $\bb$ and $\bh$, which is $\mu'_{t,l} = f_{\phi_3} (h_{t,l}, b_{t,l})$ and $\log ~ v'_{t,l} = f_{\phi_4}(h_{t,l}, b_{t,l})$. Here, we reuse the stochastic hidden states $\bh$ in the generative model in order to compress the number of the model parameters. 

{\bf{KL Annealing Trick}.} It is well known that the deep neural networks with multi-layers stochastic latent variables is difficult to train, of which one important reason is that the KL term in the loss function limited the capacity of the stochastic latent variable to compress the data information in early stages of training. The KL Annealing is a common trick to alleviate this issue. The objective function is redefined as,

\begin{equation}
\begin{aligned}
\L_\lambda(\bx) &= \E_{q_\phi(\bz|\bx)}[\sum_{t=1}^T \log~p_\theta(\bx|\bz)] \\
& \qquad - \lambda D_{KL}(q_\phi(\bz|\bx) ||p_\theta(\bz|\bx) )\\
\end{aligned}
\end{equation}

During the training process, the $\lambda$ is annealed from 0 to 1. In  previous works, researchers usually adopt the linear annealing strategy \citep{fraccaro2016sequential,goyal2017z}. In our experiment, we find that it still increases $\lambda$ too fast for Stochastic WaveNet. We propose to use cosine annealing strategy alternatively, namely the $\lambda$ is following the function $\lambda_\alpha = 1 - cos(\alpha)$, where $\alpha$ scans from $0$ to $\frac{\pi}{2}$. 

\section{Experiment}
\label{sec:experiment}

In this section, we evaluate the proposed Stochastic WaveNet on
several benchmark datasets from various domains, including natural
speech, human handwriting and human motion modeling tasks. We show
that Stochastic WaveNet, or SWaveNet in short, achieves
state-of-the-art results, and visualizes the generated samples for the
human handwriting domain. The experiment codes are publicly accessible. \footnote{https://github.com/laiguokun/SWaveNet}

\textbf{Baselines}. The following sequential generative models proposed in recent years are treated as the baseline methods:

\begin{itemize}
\item \textbf{RNN}: The original recurrent neural network with the LSTM cell. 
\item \textbf{VRNN}: The generative model with the recurrent structure proposed in \citep{chung2015recurrent}. It firstly formulates the prior distribution of $z_t$ as a conditional probability given historical data $x_{<t}$ and latent variables $z_{<t}$. 
\item \textbf{SRNN}: Proposed in \citep{fraccaro2016sequential}, and it augments the inference network by a backward RNN to better optimize the ELBO.
\item \textbf{Z-forcing}: Proposed in \citep{goyal2017z}, whose architecture is similar to SRNN, and it eases the training of the stochastic latent variables by adding auxiliary cost which forces model to use stochastic latent variables $z$ to reconstruct the future data $x_{>t}$.   
\item \textbf{WaveNet}: Proposed in \citep{van2016wavenet} and produce state-of-the-art result in the speech generation task. 
\end{itemize}

We evaluate different models by comparing the log-likelihood on the test set (RNN, WaveNet) or its lower bound (VRNN, SRNN, Z-forcing and our method). For fair comparison, a multivariate Gaussian distribution with the diagonal covariance matrix is used as the output distribution for each time stamp in all experiments. The Adam optimizer \citep{kingma2014adam} is used for all models, and the learning rate is scheduled by the cosine annealing. Following the experiment setting in \citep{fraccaro2016sequential}, we use 1 sample to approximate the variational evidence lower bound to reduce the computation cost.

\subsection{Natural Speech Modeling}

In the natural speech modeling task, we train the model to fit the log-likelihood function of the raw audio signals, following the experiment setting in \citep{fraccaro2016sequential,goyal2017z}. The raw signals, which correspond to the real-valued amplitudes, are represented as a sequence of 200-dimensional frames. Each frame is 200 consecutive samples. The preprocessing and dataset segmentation are identical to \citep{fraccaro2016sequential,goyal2017z}.  We evaluate the proposed model in the following benchmark datasets:

\begin{itemize}
\item Blizzard \citep{prahallad2013blizzard}: The Blizzard Challenge 2013, which is a text-to-speech dataset containing 300 hours of English from a single female speaker. 

\item TIMIT \footnote{https://catalog.ldc.upenn.edu/ldc93s1}: TIMIT raw audio data sets, which contains 6,300 English sentence, read by 630 speakers.
\end{itemize}

For Blizzard datasets, we report the average log-likelihood over the half-second segments of the test set. For TIMIT datasets, we report the average log-likelihood over each sequence of the test set, which is following the setting in \citep{fraccaro2016sequential,goyal2017z}. In this task, we use 5-layer SWaveNet architecture with 1024 hidden dimensions for Blizzard and 512 for TIMIT. And the dimensions of the stochastic latent variables are 100 for both datasets. 

\begin{table}[!ht]
\centering
\begin{tabu}{lrr}
\toprule
Method    & Blizzard & TIMIT \\
\midrule
RNN   
& 7413     & 26643 \\
VRNN 
& $\ge 9392$     & $\ge 28982$ \\
SRNN 
& $\ge 11991$    & $\ge 60550$ \\
Z-forcing(+kla) 
& $\ge 14226$    & $\ge 68903$ \\
Z-forcing(+kla,aux)* 
& $\ge 15024$&  $\ge  70469$ \\
\midrule
WaveNet   
& -5777  & 26074 \\
SWaveNet   &  $\ge  \textbf{15708}$        & $\ge \textbf{72463}$ \\
\rowfont{\scriptsize}           &     $(\pm 274)$ & $(\pm 639)$ \\
\bottomrule
\end{tabu}
\vspace{0.3cm}
\caption{Test set Log-likelihoods on the natural speech modeling
  task. The first group is all RNN-based models, while the second
  group is WaveNet-based models. Best results are highlighted in bold.
  $*$ denotes that the training objective is equipped with an
  auxiliary term which other methods don't have. For SWaveNet, we
  report the mean ($\pm$ standard deviation) produced by 10 different
  runs.}
\label{tab:speech}
\vspace{-0.6cm}
\end{table}

The experiment results are illustrated in Table \ref{tab:speech}. The proposed model has produced the best result for both datasets. Since the performance gap is not significant enough, we also report the variance of the proposed model performance by rerunning the model with 10 random seeds, which shows the consistence performance. Compared with the WaveNet model, the one without stochastic hidden states, SWaveNet gets a significant performance boost. Simultaneously, SWaveNet still enjoys the advantage of the parallel training compared with RNN-based stochastic models. One common concern about SWaveNet is that it may require larger hidden dimension of the stochastic latent variables than RNN based model due to its multi-layer structure. However, the total dimension of stochastic latent variables for one time stamp of SWaveNet is 500, which is twice as the number in the SRNN and the Z-forcing papers \citep{fraccaro2016sequential, goyal2017z}. We will further discuss the relationship between the number of stochastic layers and the model performance in section \ref{sec:ablation}.   

\begin{figure*}[!ht]
    \begin{minipage}{1.\textwidth}
        \centering
        \begin{minipage}[b]{0.24\columnwidth}
            \centering
            \includegraphics[width=0.95\columnwidth,height=0.5cm,clip=true]{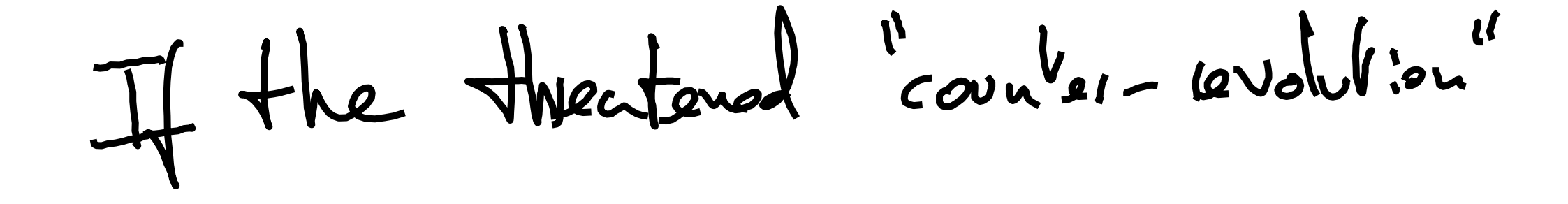}
        \end{minipage}
        \hfill
        \begin{minipage}[b]{0.24\columnwidth}
            \centering
            \includegraphics[width=0.95\columnwidth,height=0.5cm,clip=true]{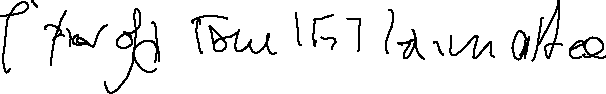}
        \end{minipage}
        \hfill
        \begin{minipage}[b]{0.24\columnwidth}
            \centering
            \includegraphics[width=0.95\columnwidth,height=0.5cm,clip=true]{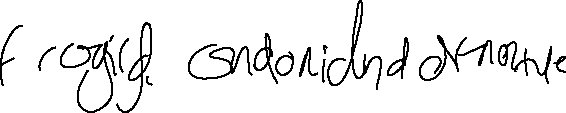}
        \end{minipage}
        \hfill
        \begin{minipage}[b]{0.24\columnwidth}
            \centering
            \includegraphics[width=0.95\columnwidth,height=0.5cm,clip=true]{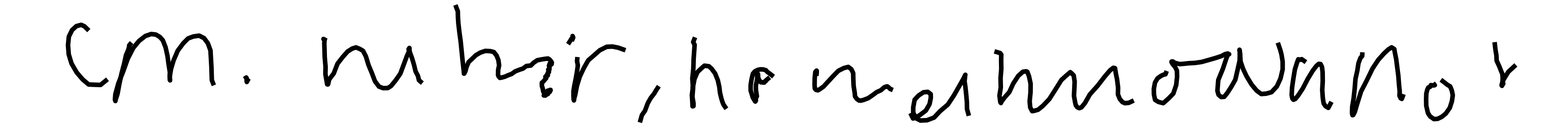}
        \end{minipage}
        \begin{minipage}[b]{0.24\columnwidth}
            \centering
            \includegraphics[width=0.95\columnwidth,height=0.5cm,clip=true]{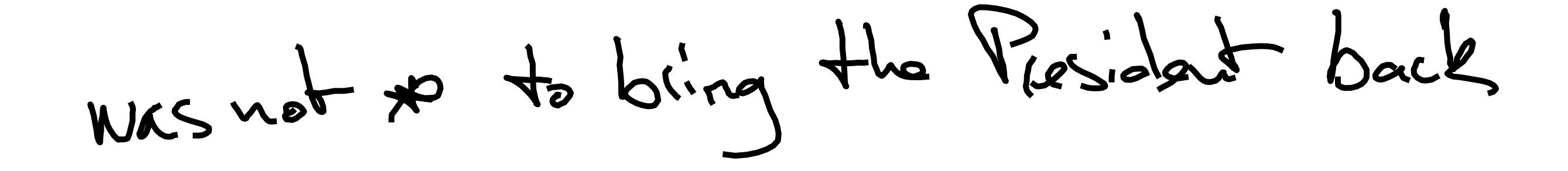}
        \end{minipage}
        \hfill
        \begin{minipage}[b]{0.24\columnwidth}
            \centering
            \includegraphics[width=0.95\columnwidth,height=0.5cm,clip=true]{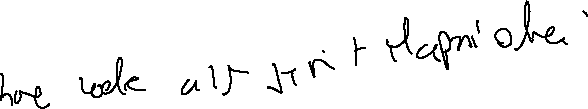}
        \end{minipage}
        \hfill
        \begin{minipage}[b]{0.24\columnwidth}
            \centering
            \includegraphics[width=0.95\columnwidth,height=0.5cm,clip=true]{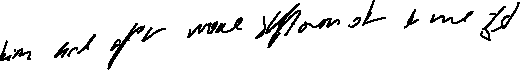}
        \end{minipage}
        \hfill
        \begin{minipage}[b]{0.24\columnwidth}
            \centering
            \includegraphics[width=0.95\columnwidth,height=0.5cm,clip=true]{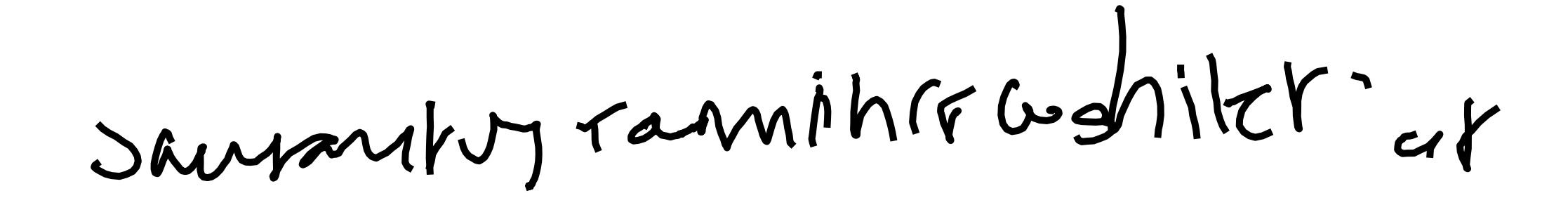}
        \end{minipage}
        \begin{minipage}[b]{0.24\columnwidth}
            \centering
            \includegraphics[width=0.95\columnwidth,height=0.5cm,clip=true]{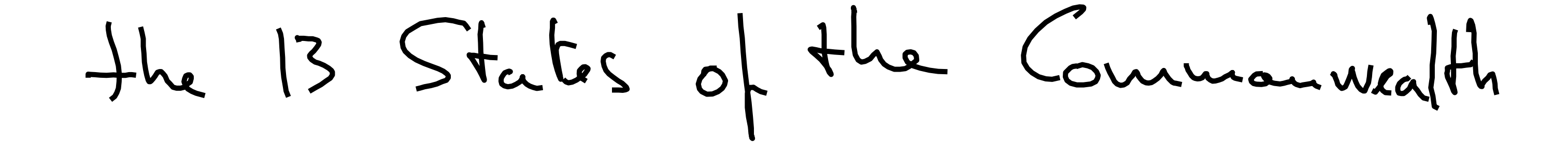}
        \end{minipage}
        \hfill
        \begin{minipage}[b]{0.24\columnwidth}
            \centering
            \includegraphics[width=0.95\columnwidth,height=0.5cm,clip=true]{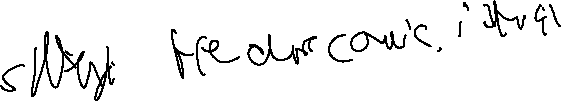}
        \end{minipage}
        \hfill
        \begin{minipage}[b]{0.24\columnwidth}
            \centering
            \includegraphics[width=0.95\columnwidth,height=0.5cm,clip=true]{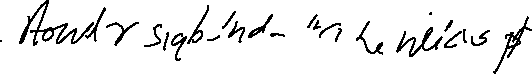}
        \end{minipage}
        \hfill
        \begin{minipage}[b]{0.24\columnwidth}
            \centering
            \includegraphics[width=0.95\columnwidth,height=0.5cm,clip=true]{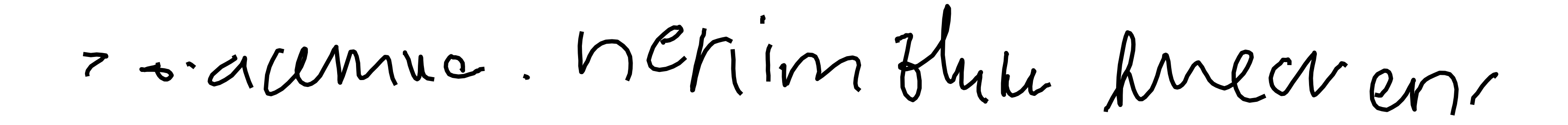}
        \end{minipage}
        \begin{minipage}[b]{0.24\columnwidth}
            \centering
            \includegraphics[width=0.95\columnwidth,height=0.5cm,clip=true]{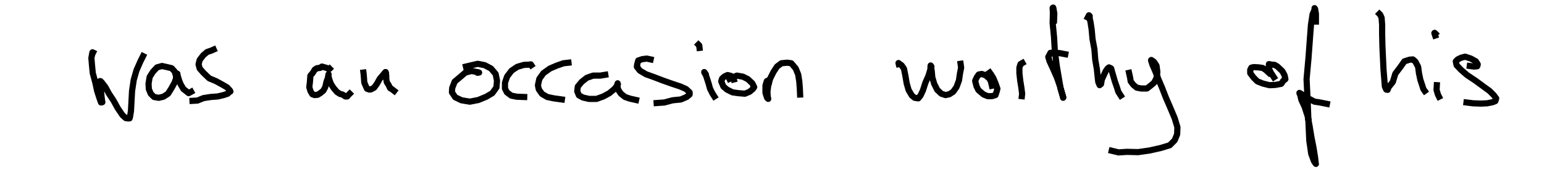}
        \end{minipage}
        \hfill
        \begin{minipage}[b]{0.24\columnwidth}
            \centering
            \includegraphics[width=0.95\columnwidth,height=0.5cm,clip=true]{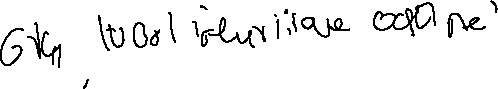}
        \end{minipage}
        \hfill
        \begin{minipage}[b]{0.24\columnwidth}
            \centering
            \includegraphics[width=0.95\columnwidth,height=0.5cm,clip=true]{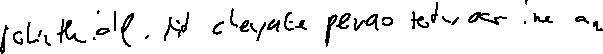}
        \end{minipage}
        \hfill
        \begin{minipage}[b]{0.24\columnwidth}
            \centering
            \includegraphics[width=0.95\columnwidth,height=0.5cm,clip=true]{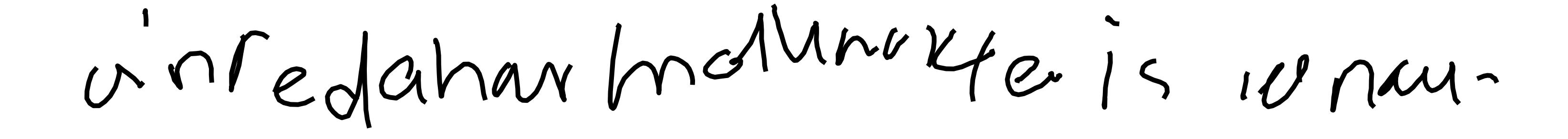}
        \end{minipage}
        \begin{minipage}[b]{0.24\columnwidth}
            \centering
            \includegraphics[width=0.95\columnwidth,height=0.5cm,clip=true]{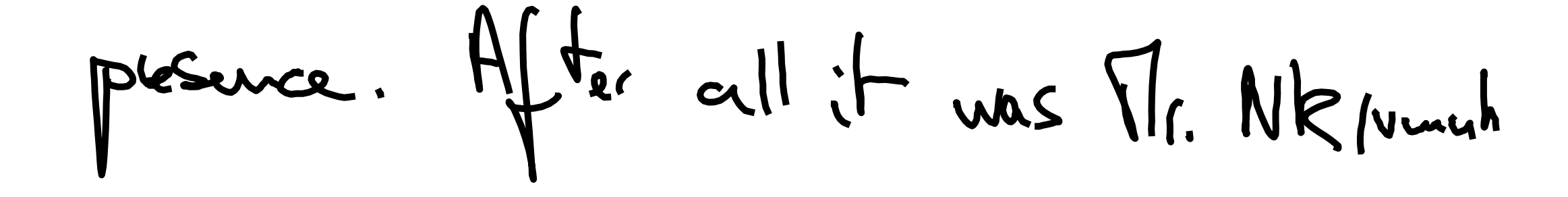}
        \end{minipage}
        \hfill
        \begin{minipage}[b]{0.24\columnwidth}
            \centering
            \includegraphics[width=0.95\columnwidth,height=0.5cm,clip=true]{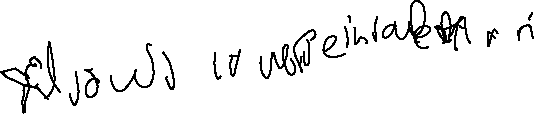}
        \end{minipage}
        \hfill
        \begin{minipage}[b]{0.24\columnwidth}
            \centering
            \includegraphics[width=0.95\columnwidth,height=0.5cm,clip=true]{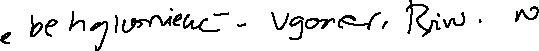}
        \end{minipage}
        \hfill
        \begin{minipage}[b]{0.24\columnwidth}
            \centering
            \includegraphics[width=0.95\columnwidth,height=0.5cm,clip=true]{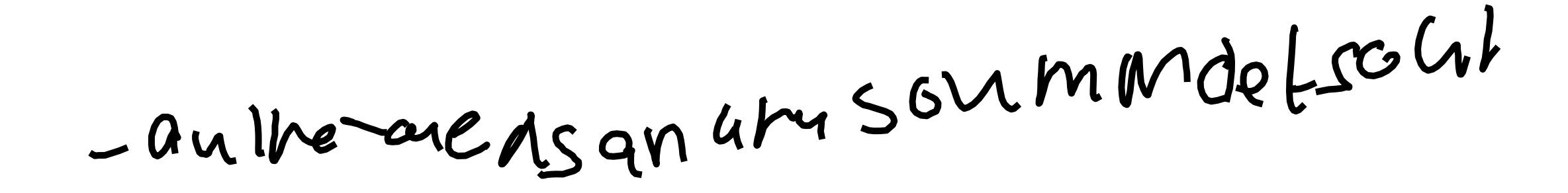}
        \end{minipage}
        \begin{minipage}[b]{0.24\columnwidth}
            \centering
            \includegraphics[width=0.95\columnwidth,height=0.5cm,clip=true]{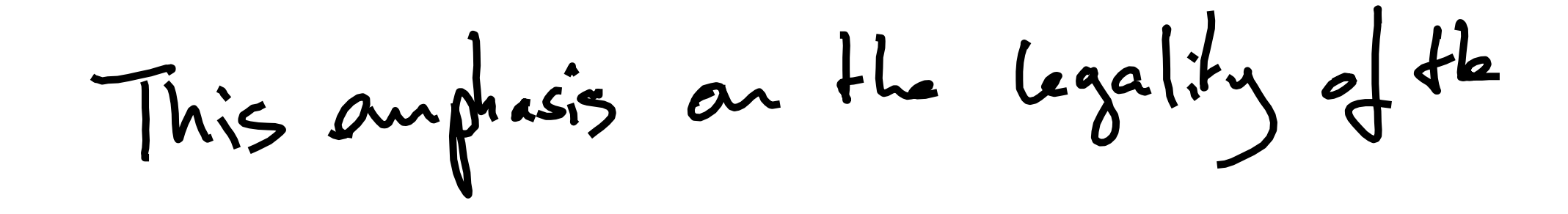}
        \end{minipage}
        \hfill
        \begin{minipage}[b]{0.24\columnwidth}
            \centering
            \includegraphics[width=0.95\columnwidth,height=0.5cm,clip=true]{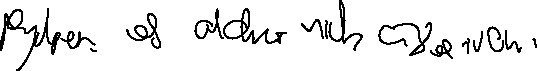}
        \end{minipage}
        \hfill
        \begin{minipage}[b]{0.24\columnwidth}
            \centering
            \includegraphics[width=0.95\columnwidth,height=0.5cm,clip=true]{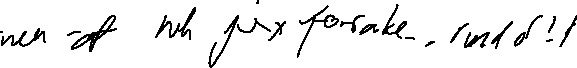}
        \end{minipage}
        \hfill
        \begin{minipage}[b]{0.24\columnwidth}
            \centering
            \includegraphics[width=0.95\columnwidth,height=0.5cm,clip=true]{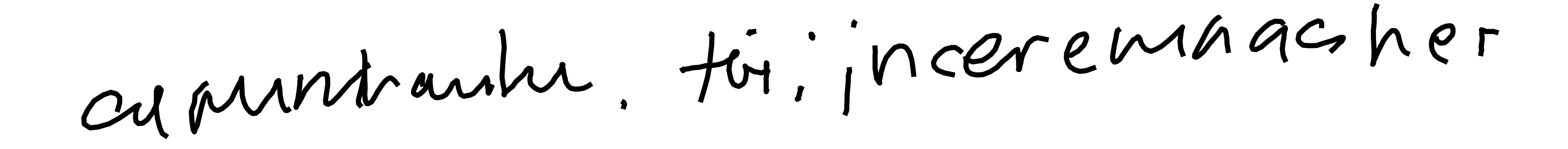}
        \end{minipage}
        \begin{minipage}{0.24\textwidth}
            \centering
            (a) Ground Truth
        \end{minipage}
        \hfill
        \begin{minipage}{0.24\textwidth}
            \centering
            (b) RNN
        \end{minipage}
        \hfill
        \begin{minipage}{0.24\textwidth}
            \centering
            (c) VRNN
        \end{minipage}
        \hfill
        \begin{minipage}{0.24\textwidth}
            \centering
            (d) SWaveNet
        \end{minipage}
    \end{minipage}
    \caption{Generated handwriting sample: (a) are the samples from
      the ground truth data. (b) (c), (d) are from RNN, VRNN and
      SWaveNet, respectively. Each line is one handwriting sample.}
    \label{fig:hwsamples}
\end{figure*}

\subsection{Handwriting and Human Motion Generation}

Next, we evaluate the proposed model by visualizing generated samples from the trained model. The domain we choose is human handwriting, whose writing tracks are described by a sequential sample points. The following dataset is used to train the generative model:

\textbf{IAM-OnDB} \citep{liwicki2005iam}: The human handwriting datasets contains 13,040 handwriting lines written by 500 writers. The writing trajectories are represented as a sequence of 3-dimension frames. Each frame is composed of two real-value numbers, which is $(x,y)$ coordinate for this sample point, and a binary number indicating whether the pen is touching the paper. The data preprocessing and division are same as \citep{graves2013generating, chung2015recurrent}.

\begin{table}[!ht]
\centering
\begin{tabular}{lr}
\toprule
Method    & IAM-OnDB \\
\midrule
RNN \citep{chung2015recurrent}    & 1016 \\
VRNN \citep{chung2015recurrent}   & $\ge \textbf{1334}$ \\
\midrule
WaveNet   &  1021 \\
SWaveNet   &  $\ge 1301$ \\
\bottomrule
\end{tabular}
\vspace{0.3cm}
\caption{Log-likelihood results on IAM-OnDB dataset. The best result are highlighted in bold.}
\label{tab:generation}
\end{table}

The quantitative results are reported in Table \ref{tab:generation}. SWaveNet achieves similar result compared with the best one, and still shows significant improvement to the vanilla WaveNet architecture. In Figure \ref{fig:hwsamples}, we plot the ground truth samples and the ones randomly generated from different models. Compared with RNN and VRNN, SWaveNet shows clearer result. It is easy to distinguish the boundary of the characters, and we can obverse that more of them are similar to the English-characters, such as ``is'' in the fourth line and ``her'' in the last line.

\subsection{Influence of Stochastic Latent Variables}
\label{sec:ablation}

The most prominent distinction between SWaveNet and RNN-based stochastic neural networks is that SWaveNet utilizes the dilated convolution layers to model multi-layer stochastic latent variables rather than one layer latent variables in the RNN models. Here, we perform the empirical study about the number of stochastic layers in SWaveNet model to demonstrate the efficiency of the design of multi-layers stochastic latent variables. The experiment is designed as follows. Firstly, we retain the total number of layers and only change the number of stochastic layers, namely the layer contains stochastic latent variables. More specifically, For a SWaveNet with $L$ layers and $S$ stochastic layers, $(S\le L)$, we eliminate the stochastic latent variables in the bottom part, which is $\{z_{1:T, l}; l < L-S\}$ in Eq.\ref{eq:forward}. 
Then for each time stamp, when the model has $D$ dimension stochastic variables in total, each layer would have $\lfloor \frac{D}{S} \rfloor$ dimension stochastic variables. In this experiment, we set $D = 500$.

\begin{figure}
\centering
\begin{subfigure}{.2\textwidth}
  \includegraphics[width=\linewidth]{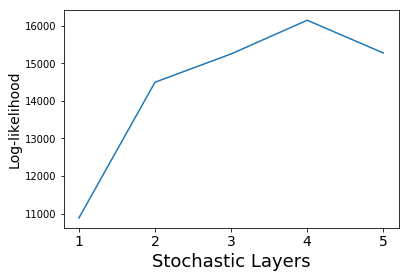}
  \caption{Blizzard}
\end{subfigure}
\begin{subfigure}{.2\textwidth}
  \includegraphics[width=\linewidth]{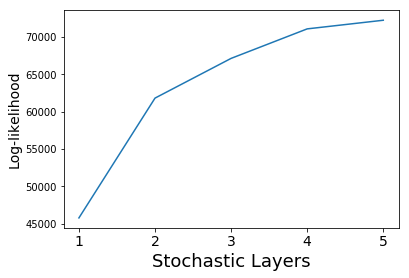}
  \caption{TIMIT}
\end{subfigure}
\begin{subfigure}{.2\textwidth}
  \includegraphics[width=\linewidth]{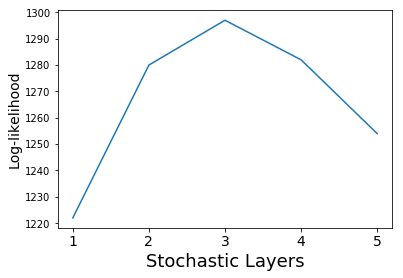}
  \caption{IAM-OnDB}
\end{subfigure}
\caption{The influence of the number of stochastic layers of SWaveNet.}
\label{fig:ablation}
\end{figure}

We plot the experiment results in Figure \ref{fig:ablation}. From the plots, we find that SWaveNet can achieve better performance with multiple stochastic layers. This demonstrates that it is helpful to encode the stochastic latent variables with a hierachical structure. And in the experiment on Blizzard and IAM-OnDB, we observe that the performance will decrease when the number of stochastic layers is large enough. Because too large number of stochastic layers would result in too small number of latent variables for a layer to memorize valuable information in different hierarchy levels.

\begin{figure}
\centering
\begin{subfigure}{.2\textwidth}
  \includegraphics[width=\linewidth]{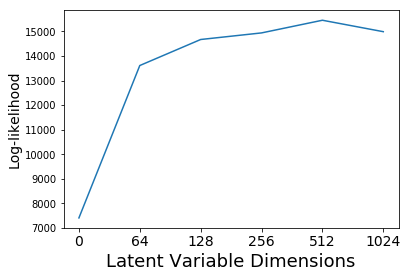}
  \caption{Blizzard}
\end{subfigure}
\begin{subfigure}{.2\textwidth}
  \includegraphics[width=\linewidth]{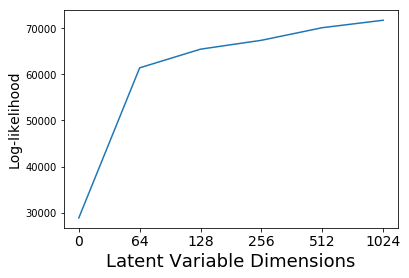}
  \caption{TIMIT}
\end{subfigure}
\begin{subfigure}{.2\textwidth}
  \includegraphics[width=\linewidth]{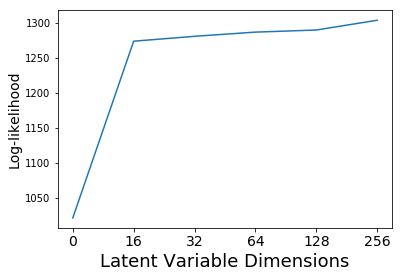}
  \caption{IAM-OnDB}
\end{subfigure}
\caption{The influence of the number of stochastic latent variables of SWaveNet.}
\label{fig:ablation-number}
\end{figure}

We also study how the model performance would be influenced by the number of stochastic latent variables. Similar to previous one, we only tune the total number of stochastic latent variables and keep rest settings unchanged, which is 4 stochastic layers. The results are plotted in Figure \ref{fig:ablation-number}. They demonstrate that Stochastic WaveNet would be benefited from even a small number of stochastic latent variables.

\section{Conclusion}
\label{sec:conclusion}

In this paper, we present a novel generative latent variable model for
sequential data, named as Stochastic WaveNet, which injects stochastic
latent variables into the hidden state of WaveNet. A new inference
network structure is designed based on the characteristic of WaveNet
architecture. Empirically results show state-of-the-art performances
on various domains by leveraging additional stochastic latent
variables. Simultaneously, the training process of WaveNet is greatly
accelerated by parallell computation compared with RNN-based
models. For future work, a potential research direction is to adopt
the advanced training strategies \citep{goyal2017z,shabanian2017variational}
designed for sequential
stochastic neural networks, to Stochastic WaveNet.

\bibliographystyle{apalike}
\bibliography{paper}

\end{document}